\documentclass[preprint,12pt]{elsarticle}

\usepackage{lineno,hyperref}
\modulolinenumbers[5]

\journal{Neurocomputing}

\usepackage[dvipsnames]{xcolor}
\usepackage{multirow}

\usepackage{amssymb}
\usepackage{latexsym}
\usepackage{algpseudocode}
\usepackage{blindtext}
\usepackage{algorithm}
\usepackage{times}
\usepackage{epsfig}
\usepackage{graphicx}
\usepackage{amsmath}
\usepackage{textcase}
\usepackage{comment}
\usepackage{breqn}
\usepackage{xcolor,balance}
\usepackage{ragged2e}
\usepackage[normalem]{ulem}
\usepackage{caption, booktabs}
\usepackage{xparse}
\usepackage{array}
\newcolumntype{L}[1]{>{\raggedright\let\newline\\\arraybackslash\hspace{0pt}}m{#1}}
\newcolumntype{C}[1]{>{\centering\let\newline\\\arraybackslash\hspace{0pt}}m{#1}}
\newcolumntype{R}[1]{>{\raggedleft\let\newline\\\arraybackslash\hspace{0pt}}m{#1}}
\newcolumntype{J}[1]{>{\justifying\let\newline\\\arraybackslash\hspace{0pt}}m{#1}}

\usepackage{enumitem}
\usepackage{float}

\begin{document}

\begin{frontmatter}




\title{Leveraging Orientation for Weakly Supervised Object Detection with Application to Firearm Localization}

\author{Javed~Iqbal}
\ead{javed.iqbal@itu.edu.pk}

\author{Muhammad~Akhtar~Munir}
\ead{akhtar.munir@itu.edu.pk}

\author{Arif~Mahmood}
\ead{arif.mahmood@itu.edu.pk}

\author{Afsheen~Rafaqat~Ali}
\ead{afsheen.57@gmail.com}

\author{Mohsen~Ali\corref{cor}}
\cortext[cor]{Corresponding author}
\ead{mohsen.ali@itu.edu.pk}

\address{Information Technology University, Lahore, 54000, Pakistan}





\begin{abstract}
Automatic detection of firearms is important for enhancing the security and safety of people, however, it is a challenging task owing to the wide variations in shape, size and appearance of firearms. 
Also, most of the generic object detectors process axis-aligned rectangular areas though, a thin and long rifle may actually cover only a small percentage of that area and the rest may contain irrelevant details suppressing the required object signatures.
To handle these challenges, we propose a weakly supervised Orientation Aware Object Detection (OAOD) algorithm which learns to detect oriented object bounding boxes (OBB)  while using Axis-Aligned Bounding Boxes (AABB) for training. The proposed OAOD is different from the existing oriented object detectors which strictly require OBB during training which may not always be present. The goal of training on AABB and detection of OBB is achieved by
employing a multistage scheme, with Stage-1 predicting the AABB and Stage-2 predicting OBB.
In-between the two stages, the oriented proposal generation module along with the object aligned RoI pooling is designed to extract features based on the predicted orientation and to make these features orientation invariant. 
A diverse and challenging dataset consisting of eleven thousand images is also proposed for firearm detection which is manually annotated for firearm classification and localization. The proposed ITU Firearm dataset (ITUF) contains a wide range of guns and rifles. The  OAOD algorithm is evaluated on the ITUF dataset and compared with current state-of-the-art object detectors, including fully supervised oriented object detectors. OAOD has outperformed both types of object detectors with a significant margin. 
The experimental results (mAP: \textit{88.3} on $\textbf{AABB}$ \&  mAP: \textit{77.5} on  $\textbf{OBB}$) demonstrate effectiveness of the proposed algorithm for firearm detection.

\end{abstract}




\end{frontmatter}


\sloppy

\section{Introduction}
\label{sec:intro}

In recent years, the world has witnessed an exponential increase in gun violence, morphing from isolated street crimes to incidences of mass shootings \cite{Howmanys35:online,Massshoo28:online,SantaFeS56:online}. Governments and private security agencies have been expanding the use of surveillance systems to monitor and secure public and private spaces. Mostly these surveillance systems are based on massive installations of camera-based surveillance systems which are mostly passive and where monitoring is delegated to the human operators. Usually, video from multiple CCTV cameras is streamed into a central station, where trained operators monitor these live footage, proactively watching for unusual activities and prohibited objects such as weapons. The operator’s ability to detect abnormality while monitoring a video feed is influenced by many variables including both technical (quality of images) and human factors such as age, experience, training  and shift duration \cite{howard2013suspiciousness}. 
\begin{figure}[t]
    \begin{center}
    \captionsetup{justification=justified}
    \includegraphics[width=0.90\linewidth]{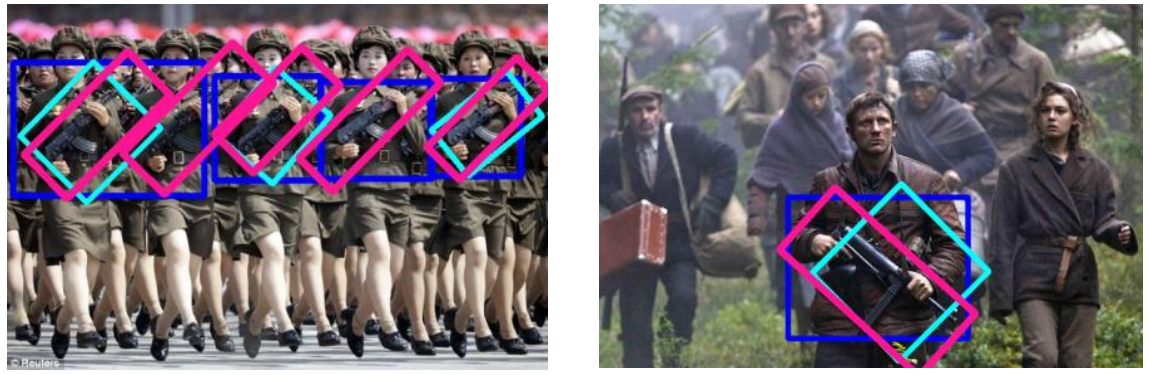}
    \caption{\small{Automatic firearms detection: Oriented bounding boxes (\textcolor{magenta}{Magenta}) detected by the proposed OAOD algorithm exhibit better accuracy and localization compared to the RoI Trans  \cite{Ding_2019_CVPR}(\textcolor{cyan}{Cyan}) and FRCNN \cite{ren2015faster}(\textcolor{blue}{Blue}) which suffer from miss-detections and poor localization. 
    }}
    \label{teaser}
    \end{center}
    \vspace{-0.5cm}
\end{figure}
Studies have shown that the human ability to detect abnormalities from live feeds reduces as the number of simultaneous feeds increases \cite{van2005cctv, sulman2008effective}. The firearm based incidents are more difficult to detect since mostly they don't involve physical altercation but just the presence of the firearm changes the dynamics of the situation. A visual firearm detection system would not only be helpful in active security monitoring but also it would be vital in monitoring harmful content on social media. Such a novel scientific solution can be embedded in surveillance systems for significant improvement in identifying potential gun violence incidence.

Despite an immense need to develop a firearms detection system, due to a number of challenges, no significant research work has yet been done in this direction. Visual firearm detection is inherently challenging due to intentional or unintentional occlusions, the close proximity of the object to the human body and design inspired for the camouflage. 
Existing visual object detectors (\cite{ren2015faster,redmon2016you,liu2016ssd}) despite being successful in detecting in a wide variety of common objects, do not perform well when dealing with firearms (Fig. \ref{teaser}). 
One of the reasons being that most of the existing object detectors predict object locations by looking at features in axis-aligned bounding boxes \cite{zhang2016faster, li2018scale, li2017perceptual, liu2019towards, ren2015faster,redmon2016you,liu2016ssd}.

\begin{figure}[t]
    \begin{center}
    \captionsetup{justification=justified}
    \includegraphics[width=0.90\linewidth]{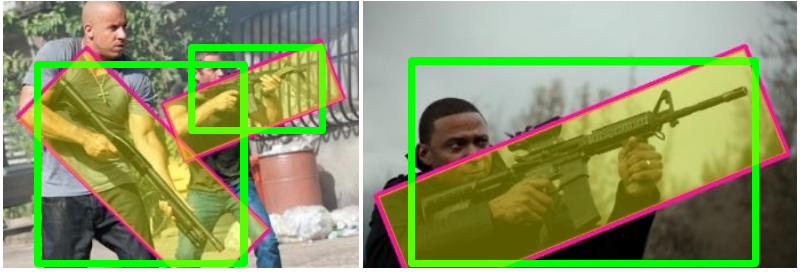}
    
        \caption{\small{Axis-Aligned Bounding Boxes (AABB) shown in \textcolor{green}{Green} are often wider than the actual object, hence features extracted from AABB are heavily affected by the background clutter. Oriented Bounding Boxes (OBB) shown in (\textcolor{magenta}{Magenta}), being aligned with the object, are relatively tighter and features pooled from OBB contain less noise from background clutter.
        }}
    \label{fig:aalign}
    \vspace{-0.5cm}
    \end{center}
\end{figure}

A physically thin and elongated structure of the rifles and small size of most guns, make these axis-aligned detectors inefficient due to low signal to noise ratio where the signal is the firearm signature and noise is everything else in the bounding box.  This problem is evident in the case of firearms being carried by a person, the axis-aligned bounding box will tend to contain substantial information belonging to the background or non-firearm objects, like the person himself (Fig.~\ref{fig:aalign}).  The inherent size \& shape variations of long and thin firearms, unfavorable viewing angles, and clutter make the detection more challenging than other objects such as human faces and vehicles.

Recently some oriented object detection methods have also been proposed that try to detect Oriented Bounding Boxes (OBB) aligned with the objects. These algorithms target applications such as ships and aeroplanes in satellite images, and text in documents, by predicting oriented region proposals \cite{ma2018arbitrary,azimi2018towards}.  However, this requires to have oriented boxes as a part of anchors at each location of the feature map, resulting in computational inefficiency due to a significant increase in the number of anchors. Also, for the training of region proposal network (RPN) and the detector itself, oriented boxes are needed as ground-truth. Annotating such boxes is quite a time consuming and erroneous, which is the reason that most datasets provide only axis-aligned boxes.  

To address these challenges, we propose an Orientation Aware multi-stage object detection system (OAOD), which is trained in a weakly supervised fashion, only on the axis-aligned bounding boxes (AABB) and orientation of the firearms.  
In our proposed system, RPN and orientation prediction are kept separate, allowing us to use a smaller number of anchors than oriented object detectors. In the first stage of our proposed system AABB and object orientation are jointly estimated, and in the second stage, a novel Oriented Proposal Generation (OPG) module is introduced to generate Oriented Region Proposal ($ORP$) by incorporating the predicted orientation information. The OPG is followed by Object Aligned Region of Interest pooling (OARoI-Pooling) to pool the features without background noise.
Our proposed system predicts both axis-aligned as well as object aligned bounding boxes, while only being trained on the axis-aligned bounding boxes and orientation information in a weakly-supervised fashion. Main contributions of the current work include:

\begin{itemize}[noitemsep,topsep=0pt]
    \item  We propose a \textbf{weakly supervised} deep learning architecture to predict Oriented Bounding Boxes (OBB)  without using OBB annotations while training.  
    \item Orientation classification and regression module are proposed to predict orientation from the axis-aligned region proposals. 
    
    \item An Oriented Proposal Generation (OPG) module is proposed to generate Oriented Region Proposals ($ORP$) followed by Object Aligned RoI-pooling (OARoI-Pooling) to pool target object features while discarding the background noise.  Such a setup results in the features that are independent of the object's orientation simplifying the task of classification and bounding box regression. Thus improving the accuracy of classifier and bounding box regressor in the last stage. 

    \item An extensive firearm dataset, ITU-Firearm (ITUF), is also proposed consisting of around 13647  annotated firearm instances in 10973 images.  
    \item Our method achieves state-of-the-art performance compared to existing methods on the proposed ITUF dataset.
\end{itemize}
For a comprehensive analysis, the proposed OAOD algorithm is compared with five existing state-of-art axis-aligned object detection methods \cite{ren2015faster, redmon2017yolo9000, redmon2018yolov3, liu2016ssd, fu2017dssd} and three oriented object detection methods \cite{Ding_2019_CVPR, yang2018r2cnn, Xia_2018_CVPR}. 
The proposed OAOD has produced an excellent performance in terms of accuracy and stability compared to these existing methods.

\section{Related Work}
\label{sec:relatedwork}
\textbf{Generic Object Detectors:} 
Significant progress has been made in developing deep-CNN based axis-aligned generic object detectors, which could be divided into two categories, including, multi-stage and single-stage detectors.
Multi-stage detectors, generally contain the first stage of RPN that selects one of the predefined anchor boxes as the proposal at each location \cite{ren2015faster, he2017mask, lin2017feature, cai2018cascade, singh2018analysis}. The next stage is to regress the final bounding boxes and classify them to the object classes.
The single-stage object detectors, like YOLO family \cite{redmon2016you,redmon2018yolov3,redmon2017yolo9000} and SSD \cite{liu2016ssd}, are well known for high detection speed but have been found to have lower performance \cite{huang2017speed}. Attributing to the class imbalance, Lin et al. \cite{lin2018focal} proposed focal loss, however, they still suffer from performance degradation while detecting small objects.
The research community is putting efforts to make object detection proposal or anchor free \cite{duan2019centernet, law2018cornernet, Tian_2019_ICCV} 
though two-stage detectors \cite{singh2018analysis, singh2018sniper, liu2018path} still have better accuracy due to better region sampling.
Axis-aligned object detectors do not handle thin and elongated objects detection challenge, where orientation makes object-size vs the AABB size disproportionate.

\begin{table}[t]
\centering
\footnotesize
\renewcommand{\arraystretch}{1.0}
\tabcolsep=3.0pt\relax
\caption{\footnotesize{An overview of different oriented object detection methods with respect to application domain, ground truth used during training and output bounding boxes (AABB or OBB).}}
\begin{tabular}{ccccc}
\hline
\multirow{2}{*}{\textbf{Methods}} & \multirow{2}{*}{\textbf{Domain}}                          & \multirow{2}{*}{\textbf{\begin{tabular}[c]{@{}c@{}}Ground Truth\\ for Training\end{tabular}}} & \multicolumn{2}{c}{\textbf{Output}}                                        \\ \cline{4-5} 
                                 &                                                           &                                                                             & \multicolumn{1}{c}{\textbf{OBB}} & \multicolumn{1}{c}{\textbf{AABB}} \\  
 \hline
RRPN \cite{ma2018arbitrary} &   Text Images & Oriented Boxes &  $\checkmark$ & 

\\ \hline

R2CNN++ \cite{yang2018r2cnn}                                                   &      Aerial Images       & Oriented Boxes & $\checkmark$ & $\checkmark$ \\ \hline

DMP-Net \cite{liu2017deep}                                                        &      Text Images  & Oriented Boxes &  $\checkmark$ &  \\ \hline

FOTS \cite{liu2018fots}                                       &               Text Images  & Oriented Boxes  & $\checkmark$ &  \\ \hline

ICN+FPN \cite{ azimi2018towards}                                                  &      Aerial Images & Oriented Boxes  & $\checkmark$ & $\checkmark$ \\ \hline

RBox-CNN \cite{koo2018rbox}                                  &               Aerial Images  & Oriented Boxes  & $\checkmark$ &  \\ \hline

\begin{tabular}[c]{@{}c@{}}RoI-Trans \cite{Ding_2019_CVPR} \end{tabular}                                   &               Aerial Images  & Oriented Boxes & $\checkmark$ &  \\ \hline
OAOD (Ours)  &  RGB Images & \begin{tabular}[c]{@{}c@{}}Axis-Aligned\\+Angle\end{tabular} & $\checkmark$ & $\checkmark$ \\ \hline
\end{tabular}
\label{papers:obb}

\end{table}

\textcolor{black}{
\textbf{Weakly Supervised Object Detectors:}
Weakly supervised object detection has been broadly studied in last few years. 
Early approaches \cite{bilen2016weakly} exploited representation learned by deep convolutional neural network pre-trained on image classification task. In general features were pooled from the regions, indicated by region proposal generation process, image classification and object detection were jointly trained on these by back-propagating the image classification loss. 
\cite{tang2017multiple}, on the other hand proposed a Multiple Instance Learning to iteratively refine the object detector by back-propagating image-level labels through multiple object detection streams. 
Based on \cite{tang2017multiple}, another work proposed is proposal cluster learning using image level annotations for object detection \cite{tang2018pcl}. This is an iterative process and assign labels on the basis of proposal clusters for refinement of instance classifier.
TS2C \cite{wei_2018_ECCV} exploits the weakly supervised object segmentation task to help the MIL based weakly supervised object detectors to concentrate on the whole object rather than just the discriminative parts. 
We present novel method using weakly supervised orientation information and axis-aligned bounding boxes for object detection with applications to firearms.
}

\textbf{Small Object Detectors:} In some small objects such as human faces contextual information may have significance for learning deep model  \cite{hu2017finding}. Similarly, features from RPN have been used for small-sized pedestrian detection \cite{zhang2016faster}.
Singh et al. proposed scale normalized training to address the problem of extreme-scale variations \cite{singh2018analysis}. 
Liu et al \cite{liu2018structure} also emphasized the significance of context and instance relationship for accurate object detection. 
However, in the case of firearms, most of the contextual objects may remain irrelevant and behave as noise by suppressing the required object information.

\begin{figure*}[t]
    \begin{center}
    \captionsetup{justification=justified}
    \includegraphics[width=1.0\linewidth]{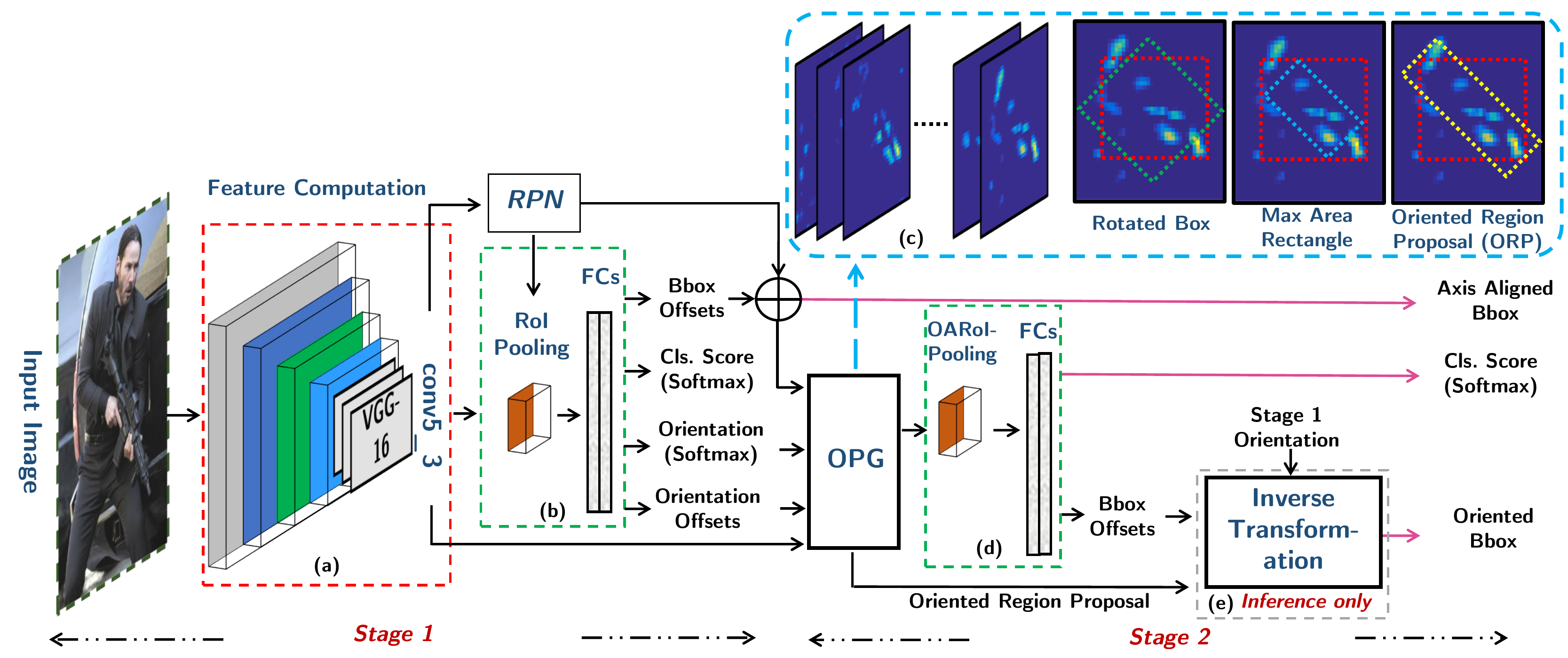}
    \vspace{-0.0cm}
    \caption{\small{
    \textbf{The architecture of proposed OAOD Algorithm:} (a) computes deep features (b) outputs object and orientation classification along with respective AABB and orientation offsets
    (c) In Stage-2, the OPG module generates Oriented Region Proposals $ORP$ using predictions from Stage-1. (d) OARoI-Pooling is applied to pool orientation independent features followed by bounding box regression (e) The final regression output are then used to generate OBB  using inverse transformation (Sec. \ref{sec:method-phase-2}).
    }}
    \label{fig:main-model}
    \end{center}
    \vspace{-0.5cm}
\end{figure*}

\textbf{Oriented Object Detectors:} 
Most of the recent oriented object detectors \cite{ma2018arbitrary, yang2018r2cnn, he2017deep, liu2018fots, azimi2018towards, Ding_2019_CVPR, koo2018rbox} are in the domains of document processing or remote sensing where objects are detected in aerial imagery (Table \ref{papers:obb}) and use OBB to train in a fully supervised way. 
OBB predicting methods try to handle challenges like dense objects, arbitrary orientation, and background noise.
Ma et al \cite{ma2018arbitrary} used the rotational formation of anchors at RPN level for text detection.
Yang et al \cite{yang2018r2cnn} used attention to improve dense objects detection in arbitrary orientations.
Ding et al \cite{Ding_2019_CVPR} proposed a rotated RoI transformer in a fully supervised way to reduce the number of anchors at RPN level. 
Nevertheless, \textcolor{black} {as indicated in Table \ref{papers:obb}},  existing oriented object detection methods use OBB information as ground truth during training. 
In contrast, we propose a cascaded approach to detect oriented objects in a weakly-supervised way, using orientation and axis-aligned bounding boxes during training.

\textbf{Firearms Detectors:} Research on visual firearm detection in images or videos is quite sparse and currently, there is no dedicated firearm detector or benchmark dataset for evaluation and comparison.
Olmos et al. used FRCNN for only handgun detection \cite{olmos2018automatic}, while no results are reported on the rifle.
Akcay et al. used FRCNN, RFCN, Yolo v2 and RCNN for gun detection in x-ray baggage security imagery \cite{akcay2018using}. 
In contrast to these existing approaches, in the current work, we propose a generic firearm detection and classification framework. 
The proposed framework is more comprehensive and does not require the OBB ground truth information during training. To the best of our knowledge, the proposed firearm detection framework is novel and has not been proposed before us.

\section{Proposed Orientation Aware Object Detector}
Most of the current object detectors predict axis-aligned bounding boxes (AABB) and for that, they analyze the features pooled from the axis-aligned window. 
Uniform pooling from an axis-aligned window may incur features containing noise due to uncorrelated background objects in the window, as shown in \ref{fig:aalign}, adversely effecting object detection performance. 
To overcome this issue, we propose an Orientation Aware Object Detector (OAOD), consisting of a cascade of two stages (\ref{fig:main-model}). 
The proposed network takes an entire image as input, localizes the firearms and simultaneously classifies them into rifles and guns.
For localization, it predicts both Oriented Bounding Boxes (OBB) and axis-aligned bounding for firearms.
Unlike other oriented object detectors \cite{ma2018arbitrary, yang2018r2cnn, liu2018fots, azimi2018towards, Ding_2019_CVPR, liu2017deep}, OAOD does not use OBB ground truth for training.
It instead relies on only the orientation information, which is much easy to annotate than OBB in the ground-truth, and learn to predict the OBB in a weakly supervised way.
In the following, both the stage-1 and the stage-2 are explained in more detail.

\subsection{OAOD Stage-1}
\label{sec:method-phase-1}
The stage-1 of OAOD consists of a Region Proposal Network (RPN) followed by a firearm localization, classification and orientation estimation network. 
\subsubsection{Region Proposal Network (RPN)}
\label{sec:method-phase-1-0}
The RPN is retrained on the firearms training dataset similar to \cite{ren2015faster}. The RPN is applied to the deep features computed by  VGG16 \cite{simonyan2014very} backbone model  to generate initial  axis-aligned region proposals, $RP_1$ which are then input to the next step.  

\subsubsection{Object and Orientation Classification Network}
\label{sec:method-phase-1-1}
During training, each region proposal $\in RP_{1}$ is associated with a unique ground-truth bounding box, based on maximum IoU between that proposal and the ground truth if maximum IoU is $\ge$ 0.50.
On the basis of this association, class label, orientation label, orientation offset, and bounding box offsets are assigned to that proposal. If maximum IoU is $<0.5$ but $\ge 0.1$, that particular proposal is labeled as background, while the others are rejected. Thus the region proposals may have classification labels as background, gun or rifle.

RoI pooling similar to FRCNN is used to pool the features to a fixed size for further processing.
These features are input to a network consisting of two fully connected layers with four separate output heads, one for each of the four tasks: object classification, orientation classification, bounding box and orientation offsets regression. 
The cross entropy loss function for firearm classification as gun, rifle, and background in stage-1 ($L^f_{1}$) is defined as:

\vspace{-0.2cm}
\begingroup
\small
\begin{align}
    L^f_{1}(p^f_{1}, u^f_{1}, n_{b}) = \sum_{i=1}^{n_{b}}\sum_{j=1}^{n_f} u^f_{1}(i,j) \text{log}(p^f_{1}(i,j)),
\label{eq4}
\end{align}
\endgroup
where $p^f_{1} \in \mathcal{R}^{n_f}$ is the predicted firearm class probability and $u^f_{1}=\{\{1,0,0\},\{0,1,0\},\{0,0,1\}\} \in \mathcal{R}^{n_f}$ is the actual firearm class label, $n_f=3$ is the number of object classes including background, gun, and rifle, and $n_{b}$ is the number of object proposals in a mini batch. 
To predict the orientation of a region proposal effectively, the objects are divided into $n_o=8$ orientation classes in the range of 0$^o$- 180$^o$ 
as shown in Fig. \ref{circle}. The other half-circle contains objects pointing in the exact opposite direction, which are also considered in the same classes as the corresponding class in the upper half-circle.  
For each region proposal the orientation classification head predicts  a label within the specified $n_o$ classes by using the
orientation loss function, $ L^{o}_{1}$ :

\vspace{-0.2cm}
\begingroup
\small
\begin{align}
    L^{o}_{1}(p^{o}_{1}, u^{o}_{1}, n_{b}) = \sum_{i=1}^{n_{b}} \sum_{j=0}^{n_o} \delta_i  u^o_1(i,j) log(p^o_1(i,j)),
\label{eq3}
\end{align}
\endgroup
where $p^{o}_{1} \in \mathcal{R}^{n_o}$ is the predicted orientation class probability and $u^{o}_{1} \in \mathcal{R}^{n_o}$ is the actual orientation class label, $n_o=8$ is the number of orientation classes, and $n_{b}$ are the number of object proposals in a mini batch corresponding to the firearms in the ground truth. 
Similarly, $\delta_i$ is an indicator variable for $i^{th}$ object proposal ensuring to ignore orientation loss during training for the background class, $\delta _{i}= 1$ if the label is gun/rifle and $0$ otherwise. 

\subsubsection{Bounding Box Regression}
\label{sec:bbox1-reg}
Alongside object classification, accurate localization is also very importance in object detection. We  train a bounding box regression head  to regress offsets. 
The objective function for the bounding box regression is given by:

\vspace{-0.2cm}
\begingroup
\small
\begin{align}
\footnotesize{
    L^b_{1}(p^b_{1}, u^b_{1}, n_{b}) = \sum_{i=1}^{n_{b}} \sum_{j=1}^4 \delta_i   \text{S}_{\ell_1} (p^b_{1}(i,j) -u^b_1(i,j))}
\label{eq1}
\end{align}
\endgroup

where $p^b_{1} = (p_x, p_y, p_w, p_h)$ are predicted bounding box offsets and $u^b_{1} = (u_x, u_y, u_w, u_h)$ are actual ground truth offsets for the respective proposal. Also, $n_{b}$ and $\delta_i$ are the same as defined above. 
The \text{S}$_{\ell_1}(\cdot)$ is smooth $\ell_1$ function

\begingroup
\small
\begin{align}
    \text{S}_{\ell_1} (x) = \left\{\begin{matrix}
0.5x^2 & \text{if~~~}   |x|<1 \\ 
|x|-0.5 & \text{Otherwise}
\end{matrix}\right.
\label{eq2}
\end{align}
\endgroup

During training, $L_1^{b}$ is back-propagated for only those object proposals which correspond to firearms in the ground truth while the others corresponding to the background are ignored by using the indicator variable $\delta_i$.  

\subsubsection{Orientation Offsets Regression}
\label{sec:orient-reg}
In addition to considering orientation as a classification task, we also rectify the predicted class mean angle (center of the bin as described in Sec. 4) based on the continuous-valued orientation ground truth ($r_{gt}$). 
\textcolor{black}{For our work, one orientation class represents degree-range :  $u^{o}_{1}-r_m$ to $u^{o}_{1}+r_m$, as described in Sec. \ref{sec:dataset}, where $r_m$ is equal to half of the bin size. In our experiments $r_m= 11.25^o$ since we have set number of classes to 8.} The orientation offset is measured as the deviation of $u^{o}_{1}$ from the ground truth $r_{gt}$. The offset  is then normalized using $r_m$ in the range of $[-1,1]$ as follows

\vspace{-0.1cm}
\begingroup
\small
\begin{flalign}
\begin{split}
    \mathcal \textit{u}_{1}^{r} = \frac{u^{o}_{1} - r_{gt}}{r_m},
    \label{eq:rorient}
\end{split}
\end{flalign}
\endgroup

\textcolor{black}{where $r_{gt}$ is subtracted from the mean angle  $u^{o}_{1}$ associated with classification task and normalized with an absolute value $r_m$ to obtain ground truth offsets for rectification of predicted mean angle class $p_{1}^o$.} At inference, offsets predicted are scaled back to the original values followed by addition to the mean angle obtained by classification. This helps in better localizing the oriented area to pool features, removing background noise and clutter more effectively than procedure followed in Sec \ref{sec:method-phase-1-1}. The loss function for orientation offsets regression is as follows

\begingroup
\small
\begin{align}
\footnotesize{
    L^r_{1}(p^r_{1}, u^r_{1}, n_{b}) = \sum_{i=1}^{n_{b}} \sum_{j=1}^{n_o} \delta_i   \text{S}_{\ell_1} (p^r_{1}(i,j) -u^r_{1}(i,j))},
\label{eq:rorient-loss}
\end{align}
\endgroup

where $p^r_{1}$ represents the regressed orientation offsets and $u^r_{1}$ shows orientation ground truth offsets, $n_{b}$ and $\delta_i$ are the same as defined above. The \text{S}$_{\ell_1}(\cdot)$ is smooth $\ell_1$ as defined in \eqref{eq2}.

The overall training objective function for OAOD stage-1 is a weighted combination of the individual losses of object and orientation classification along with bounding box and orientation offsets regression 

\begingroup
\small
\begin{flalign}
\begin{split}
    {L}_{1} = \alpha L^f_{1}(p^f_{1}, u^f_{1}, n_{b}) +  \beta L^{o}_{1} (p^{o}_{1}, u^{o}_{1}, n_{b})\\ + 
     \gamma L^b_{1}(p^b_{1}, u^b_{1}, n_{b}) + \eta L^r_{1}(p^r_{1}, u^r_{1}, n_{b})
    \label{eq5}
\end{split}
\end{flalign}
\endgroup

where $\alpha$, $\beta$, $\gamma$, and $\eta$  are normalization weights to assign relative importance to each term in the objective function.  The bounding box offset regression targets $u^b_{1}$ are also normalized within the same range of [-1,+1].

\subsection{OAOD Stage-2}
\label{sec:method-phase-2}
Output of the stage-1 is offsets for the Axes Aligned Bounding Boxes (AABB), their orientation class, orientation offset, and firearm classification result. 
In order to remove the noisy features belonging to the background, we use the output of the stage-1 to generate  Oriented Region Proposals ($ORP$) by Oriented Proposal Generation  (OPG) module, perform OARoI-Pooling for these proposals before presenting to the stage-2 classifier and regressor that generates Oriented Bounding Boxes (OBB). 
These steps are discussed in more detail in the following sections.

\subsubsection{Updating Region Proposals for Stage-2}

The bounding box offsets $p^b_{1}$, output by stage-1, are used to translate and scale region-proposals $RP_1$ to get $RP_2$,  $RP_2=RP_1+p^b_1$.  
Therefore,  $RP_2$ IoU with the corresponding ground truth bounding box $u^b_1$ may get changed requiring a revision of labels and training offsets for this stage. 
New ground-truth labels (${u^f_{2}, u^b_{2}}$) for each updated region proposal are recomputed by considering its IoU with the ground-truth bounding boxes.

\subsubsection{Oriented Proposal Generation}
\begin{figure}[t]
    \begin{center}
    \captionsetup{justification=justified}
    \includegraphics[width=0.9\linewidth]{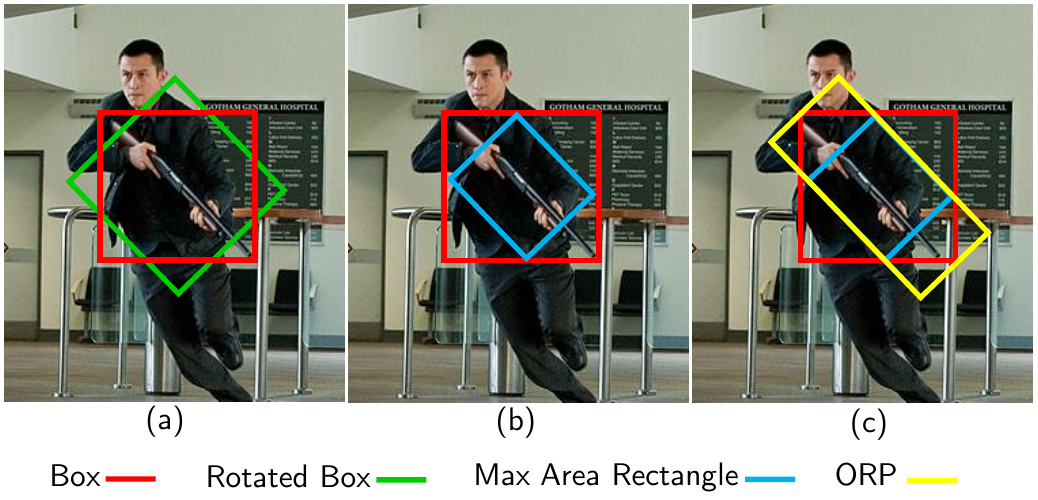}
    \caption{\small{Oriented Proposal Generation  process (OPG): (a) AABB and its rotated version using angle information, (b) Maximum area oriented rectangle inscribed in the actual box and (c) $ORP$, used in second stage for OARoI-Pooling.
    }}
    \label{figcomp:maxarea}
    \end{center}
    \vspace{-0.5cm}
\end{figure}

\textcolor{black}{
For $i_{th}$ region proposal, adding the  predicted offset $p_1^r(i)$ to the predicted mean angle $p_1^{o}(i)$ of the orientation class, we compute the updated orientation $\theta_{i}$, where $\theta_i=p^{o}_{1} (i)+r_m \times p_1^r(i)$, and $r_m=11.25^{o}$ in our case (Sec. \ref{sec:orient-reg}).} 
This updated angle is then used to generate Oriented Region Proposals ($ORP$), aligned with the firearm object.
If $RP_2$ is directly rotated using $\theta_i$, it gets aligned with the firearm but it is not compact and may encapsulate even more background information than the original $RP_2$.
To address this issue a maximum area oriented rectangle is computed inside  $RP_{2}$ such that the longitudinal axis of this rectangle is aligned with $\theta_i$  as shown in Fig. \ref{figcomp:maxarea}. 
This oriented rectangle removes the extra background information, however, in many cases, it does not cover the full length of the object. Therefore, to obtain $ORP$, the maximum area rectangle is extended along the longitudinal axis till the corners of $RP_2$. As shown in Fig. \ref{figcomp:maxarea}(c), the $ORP$ is aligned with the axis of the object and is relatively tighter than both $RP_2$ and its rotated version (green rotated box in Fig. \ref{figcomp:maxarea} (a)).

\subsubsection{Object Aligned RoI-Pooling}
Since $ORP$ consists of rectangles that are not axis-aligned, the pooling algorithm is modified. 
\textcolor{black}{We define an Object Aligned RoI-Pooling (OARoI-Pooling) process to pool values from the ORP. Unlike RoI pooling in stage-1, the OARoI-Pooling pools values from an oriented grid instead of an axis-aligned grid where the oriented grid is generated over the ORP.
}
The oriented feature map resulted from OARoI-Pooling is used for the final classification of the object and the bounding box regression \textcolor{black}{(see Fig. 3 (d))}. 
It should be noted that after OARoI-Pooling, the pooled values become invariant to the orientation of the object in $RP_1$ thus making  easier for the classifier to perform prediction.

\subsubsection{Oriented Object Detection}
An oriented object detection sub-network is trained to predict the classification score and bounding box offsets over OARoI-pooled features. 
These offsets are then applied to the $ORP$ before further processing.
The design of oriented object detection sub-network layers is similar to the stage-1 object classification and bounding box regression layers. 
The objective function for this sub-network consists of two losses including oriented object classification loss and bounding box regression loss . The cross entropy loss $L^f_{2}$ is given by:

\begingroup
\small
\begin{align}
    L^f_{2}(p^f_{2}, u^b_{2}, n_{b}) = \sum_{i=1}^{n_{b}}\sum_{j=1}^{n_f} u^f_{2}(i,j) \text{log}(p^f_{2}(i,j))
\label{eq400}
\end{align}
\endgroup

where $p^f_{2} \in \mathcal{R}^{n_f}$ is the predicted firearm class probability and $u^f_{2} \in \mathcal{R}^{n_f}$ is the updated firearm class label, while the rest of parameters are similar to \eqref{eq4}.
The bounding box regression loss $L^b_{2}$ is as follows:

\begingroup
\small
\begin{align}
    L^b_{2}(p^b_{2}, u^b_{2}, n_{b}) = \sum_{i=1}^{n_{b}} \sum_{j=1}^4 {\Theta_i} \text{S}_{\ell_1} (p^b_{2}(i,j) -u^b_{2}(i,j))
\label{eq100}
\end{align}
\endgroup

\begin{algorithm}[t]
\textbf{Input:} \small{$\theta=\{\theta_i\}_{i=1}^{n_p}$, $\textbf{x}=\{x_\min^i,y_\min^i,x_\max^i,y_\max^i\}_{i=1}^{n_p}$ where $\textbf{x}=RP_{2}$ and,
\textbf{u}=\{$u_\min^i,v_\min^i, u_\max^i,v_\max^i\}_{i=1}^{n_p}$ are the corresponding input \& output boxes from stage-2 respectively, $n_p$: total number of object proposals} \\ 
\textbf{Output:} Oriented Bounding Boxes $B_o=\{B^i_o\}_{i=1}^{n_p}$
    \caption{Inverse Transformation}
    \begin{algorithmic}[1]

    \vspace{0.1cm}
     \For{\textit{$i \leftarrow 1:n_p$}}

     \vspace{0.2cm}
     \State $C_x^i \leftarrow (x_\min^i + x_\max^i) / 2$ ,  $C_y^i \leftarrow (y_\min^i + y_\max^i) / 2$

    \vspace{0.2cm}
    \small{\State $T^i=\begin{bmatrix} \cos{\theta}_i & \sin{\theta}_i & C_x^i\cos{\theta}_i+C_y^i\sin{\theta}_i-C_x^i  \\ -\sin{\theta}_i & \cos{\theta}_i & -C_x^i\sin{\theta}_i+C_y^i\cos{\theta}_i-C_y^i \\ 0 & 0 & 1 \end{bmatrix}$}
    \vspace{0.2cm}
    \State $M_{c}^i=\begin{bmatrix} u_\min^i & u_\max^i & u_\min^i & u_\max^i \\ v_\min^i & v_\min^i & v_\max^i & v_\max^i \\ 1 & 1 & 1 & 1 \end{bmatrix}$
    \vspace{0.2cm}
    \State $B^i_o \leftarrow T^i M_{c}^i$
     \EndFor
     \end{algorithmic}
\end{algorithm}

where $\Theta_i$ is an indicator variable such that, $\Theta_i=1$ if orientation is $0^o$ or $90^o$ and $\Theta_i=0$ otherwise. Hence, loss is backpropagated only if the firearm is vertically or horizontally axis-aligned. 
For these two angles, the $RP_{2}$ and the $ORP$ remain the same; hence ground truth boxes could be used to train the stage-2 bounding box regression. 
The combined objective function for this oriented object detection sub-network is given below: 

\begingroup
\small
\begin{flalign}
\begin{split}
    L_{2} =  L^f_{2}(p^f_{2}, u^f_{2}, n_{b}) + L^b_{2}(p^b_{2}, u^b_{2}, n_{b})
    \label{eq6}
\end{split}
\end{flalign}
\endgroup

\subsubsection{Oriented Bounding Boxes Output}
The bounding box offsets, $p_2^b$, predicted in stage-2 are used to update $ORP$. An inverse transformation is constructed to map this adjusted $ORP$ back to the original image,  using the orientation $\theta_i$ and the $RP_{2}$ center positions from stage-2. 
The output of this final step gives us OBB. 
The step-wise details are provided in Algorithm-1. 

Note that: during the inference time, we use bounding box output by stage-1 as the AABB, and OBB generated by stage-2. However, in both cases, we use the classification probability from stage-2. 

\vspace{-0.1cm}
\section{ITU Firearms Dataset (ITUF)}
\label{sec:dataset}
We have collected a large dataset of images containing firearms, named as ITUF. Axis-aligned bounding box (AABB) of each firearm in each image has been hand-annotated.  Dataset has been divided into training and testing splits, for the testing split  OBB were also manually annotated to enable comparison with existing OBB predicting algorithms. As per our knowledge, ITUF is the first large firearm dataset in the public domain. ITUF captures varied scenes (indoor, outdoor, lighting conditions) \& scenarios (firearms pointed, carried, lying on tables/ground/racks) and contains various makes and models of firearms (from pistols to AK-47). 
This diversity makes ITUF a challenging and realistic dataset for the firearm detection task. 
\begin{figure}[ht]
    \begin{center}
    \captionsetup{justification=justified}
    \includegraphics[width=0.8\linewidth]{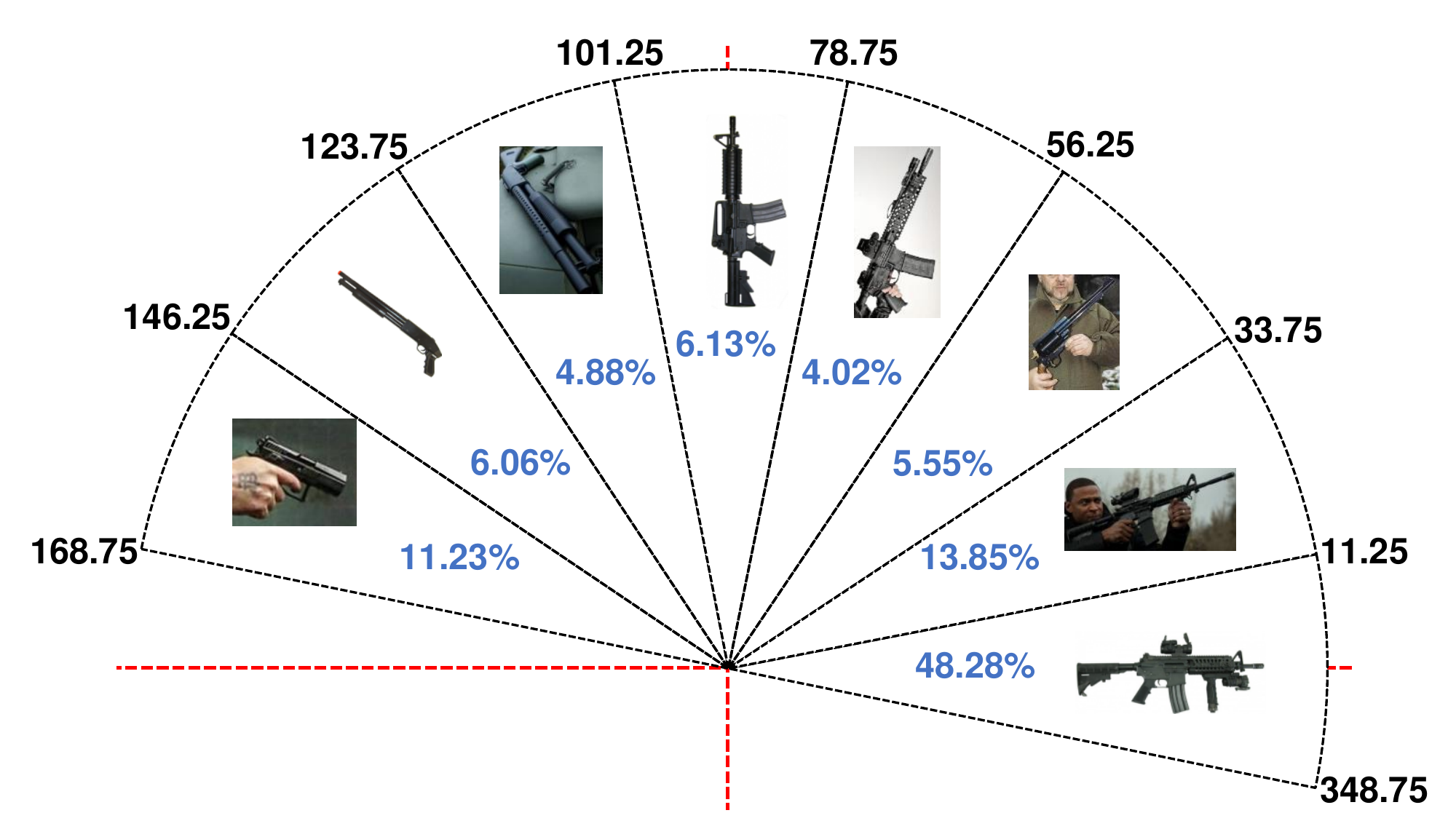}
    \caption{\small{ Orientation is divided in 8 classes considering firearms oriented $\theta+180^o$ in the same class as the firearms with $\theta$ class.  
     Values in \textcolor{blue}{blue} show the Firearms distribution over different $\theta$ classes.}}
    \label{circle}
    \end{center}
\vspace{-0.4cm}
\end{figure}

\textbf{Data Collection and Annotation: } ITUF was collected from the web by incorporating keywords, such as weapons, wars, pistol, movie names, firearms, types of firearms, sniper, shooter, corps, guns and rifles, in the web search. Results were cleaned to remove images not containing firearms, duplicates and synthetic ones.
The final dataset consists of $10,973$ fully annotated images with 13647 firearm instances. 

We have divided firearms into two classes; `Gun' class includes different variations of pistols and revolvers; whereas `Rifle' class contains hunting-rifles to AK-47 (including small machine guns). AABB for each firearm in every image is tagged by an annotator, along with a class label and an angle representing the orientation of the firearm.
Orientation is annotated as the angle made by line joining muzzle and the back tip (hammer or butt) of the firearm. 
Orientations are quantized into 8 bins as shown in Fig. \ref{circle}, and each bin is treated as a class with the value equal to the center of the bin. Each class also includes orientations which are 180$^o$ flipped versions of the angles shown in Fig. \ref{circle}. For example, class 0 spans 348.75$^o$ to +11.25$^o$ as well as 168.75$^o$ to 191.25$^o$. 
\textcolor{black}{The orientation class associated to each firearm represent the center of the associated quantization bin, and is named as \textbf{mean angle}.}

We believe that this dataset will help the researchers to develop algorithms for firearm detection not just for security but also for the multi-media content analysis, including AR and VR environments as well. It will also help media and content distribution companies to categorize what content is feasible for age-appropriate consumption. More dataset details may be found in the supplementary material and at the project homepage\footnote{http://im.itu.edu.pk/orientation-aware-firearms-detection/}.

\section{Experiments and Results}

The proposed OAOD algorithm is trained and evaluated on the ITUF dataset and is compared with current state-of-the-art axis-aligned and oriented object detection algorithms. 
A thorough ablation study is performed to validate different parameters and aspects of the proposed OAOD algorithm.

\subsection{Experimental Setup}
\label{sec:exp-setup}
In firearm detection experiments, we localize each firearm in an input image and classify it as a rifle or a gun. For the comparison with the axis-aligned object detection methods, AABB from stage-1 and classification score from stage-2 are used, whereas the output of stage-2 is used for comparisons with OBB detection methods.  OAOD is only trained on AABB and orientation ground-truth information, while OBB are not used for training. 

\noindent \textbf{Implementation Details:} High-resolution images in ITUF are resized to a shorter side of 480 or larger side of 800 pixels preserving the aspect ratio.
Due to limited GPU memory, a single image per batch is processed. 
The initial learning rate and momentum are set to 0.001 and 0.90 respectively.
A weight decay of 0.0005 with SGD optimizer is used. 
VGG16 pre-trained on Imagenet \cite{simonyan2014very}, is used as a backbone network.
Caffe is used as an implementation framework and training is performed on a single core-i5 machine with 32GB RAM and a GTX 1080 GPU with 8GB memory.
The hyper parameters in \eqref{eq5},  $\alpha$, $\gamma$  \& $\eta$ are set to 1.0. The parameter $\beta$ is set to 0.1, by validating over a wide range in search of optimal value (Sec. \ref{sec:ablation}).

\begin{table}[h]
\footnotesize
\centering
\renewcommand{\arraystretch}{1.3}
\tabcolsep=1.9pt\relax
\caption{\footnotesize{OAOD-AA  (no orientation offsets) \& OAOD-AA+ (with orientation offsets) vs state-of-the-art AABB object detectors at multiple IoU levels. $AP_g$ \& $AP_r$ : Average Precision of gun \& rifle respectively. Highest values shown in \textcolor{red}{Red}, $2^{nd}$ highest shown in  \textcolor{blue}{Blue}.}}
\begin{tabular}{c|ccc|ccc|ccc}
\hline
\multirow{2}{*}{\begin{tabular}[c]{@{}c@{}}  Methods \end{tabular}}                                     & \multicolumn{3}{c|}{\text{$AP_{40}$}}                & \multicolumn{3}{c|}{\text{$AP_{50}$}}                & \multicolumn{3}{c}{\text{$AP_{60}$}}  \\ \cline{2-10} 
                                                      & \text{$AP_g$}   & \text{$AP_r$}   & \text{$mAP$}   & \text{$AP_g$}   & \text{$AP_r$}   & \text{$mAP$}   & \text{$AP_g$}   & \text{$AP_r$}   & \text{$mAP$}  \\ \hline
\begin{tabular}[c]{@{}c@{}}YOLOv2 \end{tabular}   & 70.7          & 83.3          & 77.0           & 62.3          & 77.0           & 69.6          & 41.9          & 62.9          & 52.4 \\ 
YOLOv3                                               & 80.8          & 78.6          & 79.8          & 76.0          & 70.7          & 73.4          & 64.3          & 59.0          & 61.7 \\ 

SSD                                               & 70.6          & 79.0          & 74.8          & 65.6          & 73.0          & 69.3          & 55.2          & 58.2          & 56.7 \\ 

DSSD                                               & 77.4          & 78.9          & 78.1          & 73.0          & 72.3          & 72.7          & 63.2          & 58.9          & 61.1  \\ 

\begin{tabular}[c]{@{}c@{}}FRCNN\end{tabular} & \textcolor{black}{88.7} & \textcolor{black}{89.0}    & \textcolor{black}{88.9}    & \textcolor{black}{80.2} & \textcolor{black}{79.4} & \textcolor{black}{79.8}    & \textcolor{blue}{67.8} & \textcolor{black}{68.3} & \textcolor{black}{68.1}\\ 
\hline

\textbf{OAOD-AA}                                         & \textcolor{blue}{88.8}    & \textcolor{blue}{89.6} & \text{\textcolor{blue}{89.2}} & \textcolor{blue}{84.4} & \textcolor{blue}{86.4} & \text{\textcolor{blue}{85.4}} & \textcolor{black}{67.0}    & \textcolor{blue}{74.0} & \text{\textcolor{blue}{70.3}}  \\

\textbf{OAOD-AA+}                                         & \textcolor{red}{89.6}    & \textcolor{red}{90.2} & \text{\textcolor{red}{89.9}} & \textcolor{red}{87.6} & \textcolor{red}{88.9} & \text{\textcolor{red}{88.3}} & \textcolor{red}{73.2}    & \textcolor{red}{78.1} & \text{\textcolor{red}{75.7}}  \\ \hline

\end{tabular}

\label{tab:ioumap-axis}
\end{table}

\begin{figure*}[!htb]
    \begin{center}
    \captionsetup{justification=justified}
        \includegraphics[width=\linewidth]{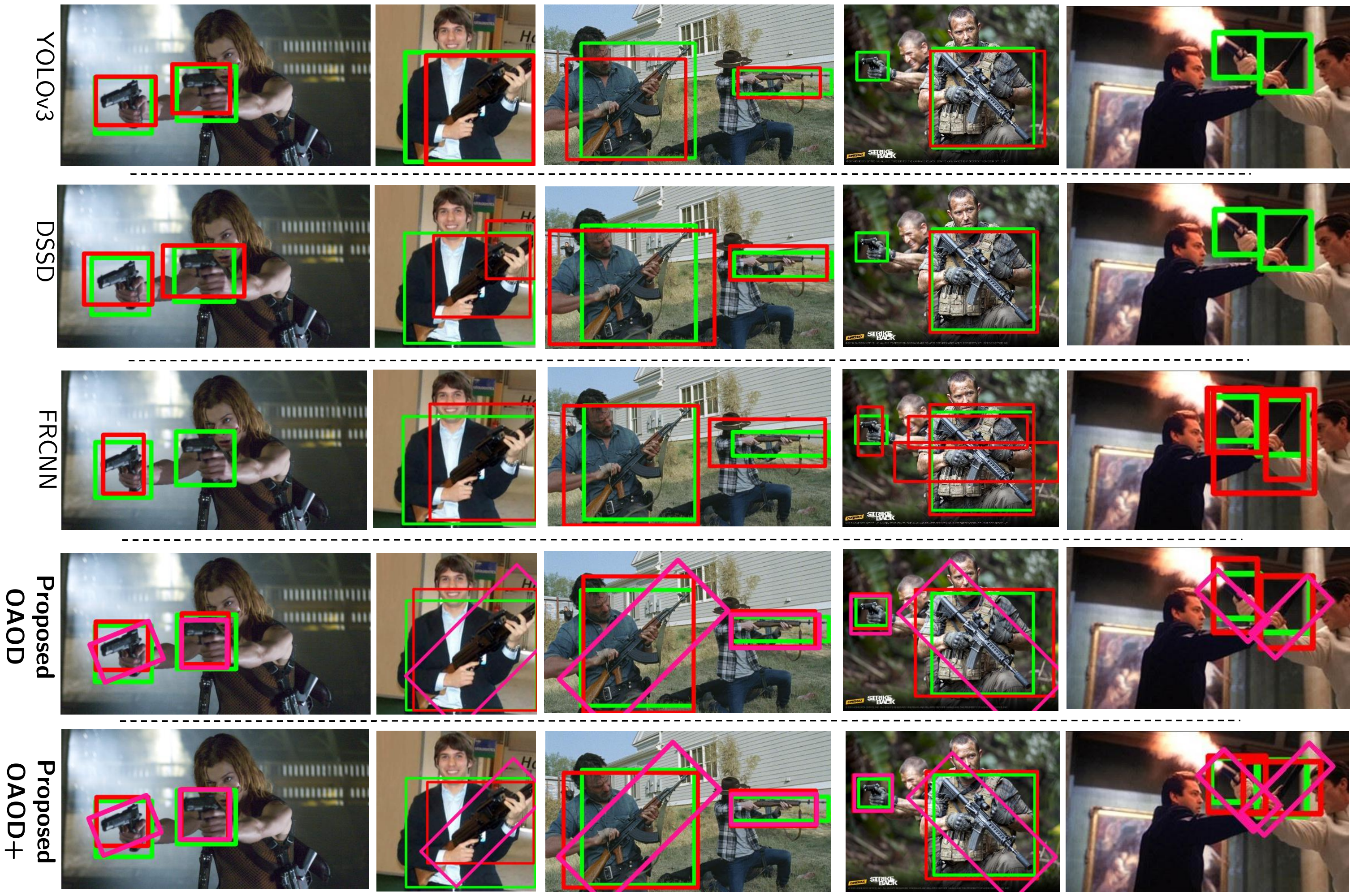}
        \caption{\small{Qualitative comparison of OAOD with current state-of-the-art object detectors, including YOLOv3, DSSD and FRCNN  at IoU=0.50. \textcolor{green}{Green}: ground truth, \textcolor{red}{Red}: AABB detections of respective algorithms. Our proposed approach results in reduced miss-detections, accurate localization and lower false detections. The \textcolor{magenta}{Magenta} boxes in the last two rows are OBB detections of OAOD algorithm, one with only orientation classification (OAOD), and last one (OAOD$+$) with orientation offset regression. OAOD$+$ results in boxes tightly bounding the object.
        }}
        \label{compresults}
    \end{center}
    \vspace{-0.5cm}
\end{figure*}


\noindent \textbf{Training Scheme: }
We train the two OAOD stages one by one. Initially stage-1 (Fig. \ref{fig:main-model}) is trained to predict the AABB, classification score and orientation information with loss function ${L}_{1}$ (\ref{eq5}). 
After a sufficient number of epochs, we train stage-2 along with fully connected layers of stage-1. The bounding boxes and the orientation information predicted by stage-1 are passed as input to stage-2 as described in Sec. \ref{sec:method-phase-2}. 
$L_b^{s2}$ in stage-2 is incorporated only if the orientation for the region proposal is 0$^o$ or 90$^o$, while the classification loss is used for every instance as given by \eqref{eq6}.
To avoid over-fitting, a dropout of 0.5 is used between the fully connected layers during training.

\subsection{Comparison with Existing State-of-the-art Techniques}
The trained OAOD and other state of the art algorithms are evaluated over the ITUF test set for both AABB and OBB predictions. 
To understand the effect of orientation prediction we present results on the both  sub-tasks, orientation classification (see Sec. \ref{sec:method-phase-1-1}) and orientation regression (see Sec. \ref{sec:orient-reg}).
OAOD pipelines only with orientation classification are named OAOD-AA \& OAOD-OB (as the AABB and OBB predictions respectively), whereas  OAOD-AA+ \& OAOD-OB+ are used for regression, that is with orientation offsets added in the pipeline (see Fig. \ref{fig:main-model}).
Our model (with orientation regression) produces mAP of 88.3\% and 77.5\% (at IoU=0.50) for AABB and OBB respectively. 
Due to $ORP$ offsets regression and stage-2 classification, the proposed OAOD avoids miss-detection and multiple detections while performing more accurate localization. 
Employing a deeper backbone network such as ResNet-101 \cite{he2016deep} may result in further improved accuracy at the cost of increased space complexity.

\subsubsection{Comparison with Axis-Aligned Bounding Box Methods}
We compare OAOD against current state-of-the-art one-stage and two-stage AABB object detection algorithms such as SSD, DSSD, YOLOv2, YOLOv3 and FRCNN. 
These methods were trained on the same ITUF training dataset. All parameters in these algorithms were set as recommended by the original authors.
\vspace{-0.0cm}

The proposed OAOD has outperformed the compared methods by achieving better mAP compared to both single-stage and multi-stage detectors as shown in Table \ref{tab:ioumap-axis}.
This is attributed to $ORP$ generated by the OPG module with OARoI-Pooling which removes much of the noisy features related to the background, making the stage-2 more accurate.
Secondly, OAOD remains stable (Table \ref{tab:ioumap-axis}) 
as IoU levels are varied, despite the fact that the model is trained for IoU=0.50 only. 
Specifically compared to the baseline FRCNN, our proposed OAOD-AA and OAOD-AA+ have achieved increased mAP by 6.6\% and 9.6\%  respectively, at IoU=0.50. Compared to the single-stage axis-aligned object detectors, the OAOD-AA and OAOD-AA+ improve the performance by a minimum of 14.0\% and 16.9\% respectively.
The qualitative results are presented in Fig. \ref{compresults}. The proposed OAOD performs excellently by avoiding miss detections and produces better localization.

\begin{table}[t]
\footnotesize
\centering
\renewcommand{\arraystretch}{1.5}
\tabcolsep=2.9pt\relax
\caption{\footnotesize{Comparison of the proposed OAOD-OB (no orientation offsets) \& OAOD-OB+ (with orientation offsets) with state-of-the-art OBB detectors at different IoU levels.
$OBB_{rot}$ are rotated version of AABB whereas $OBB_{ann}$ are manually annotated oriented boxes. Highest values shown in \textcolor{red}{Red}, $2^{nd}$ highest shown in  \textcolor{blue}{Blue}
}}

\begin{tabular}{c|c|cc|cc}
\hline
\multirow{2}{*}{Methods} & \multirow{2}{*}{Baseline} & \multicolumn{2}{c|}{$OBB_{rot}$}   & \multicolumn{2}{c}{$OBB_{ann}$} \\
\cline{3-6}
           &  & \text{$AP_{50}$}    & \text{$AP_{60}$}    & \text{$AP_{50}$} & \text{$AP_{60}$} \\ \hline
R2CNN$++$            & ResNet-50   & 54.5          & 44.8          & 43.0          & 27.9      \\
DOTA-FRCNN                 & ResNet-101 & 58.7          & \textcolor{black}{46.9}  & 57.1          & 46.3    \\
RoI-Trans            & ResNet-101    & \textcolor{black}{77.5}    & 45.9    & \textcolor{black}{68.5}  & \textcolor{black}{48.5}   \\
\hline
OAOD-OB            & VGG16 & \textcolor{blue}{77.9} & \textcolor{blue}{62.2} & \textcolor{blue}{69.7} & \textcolor{red}{50.2} \\
OAOD-OB+            & VGG16     & \textcolor{red}{82.3} & \textcolor{red}{63.8} & \textcolor{red}{77.5} & \textcolor{blue}{49.6} \\
\hline
\end{tabular}

\label{tab:oriented-comparison}
\end{table}

\begin{figure*}[!htb]
    \begin{center}
        \includegraphics[width=0.9\linewidth]{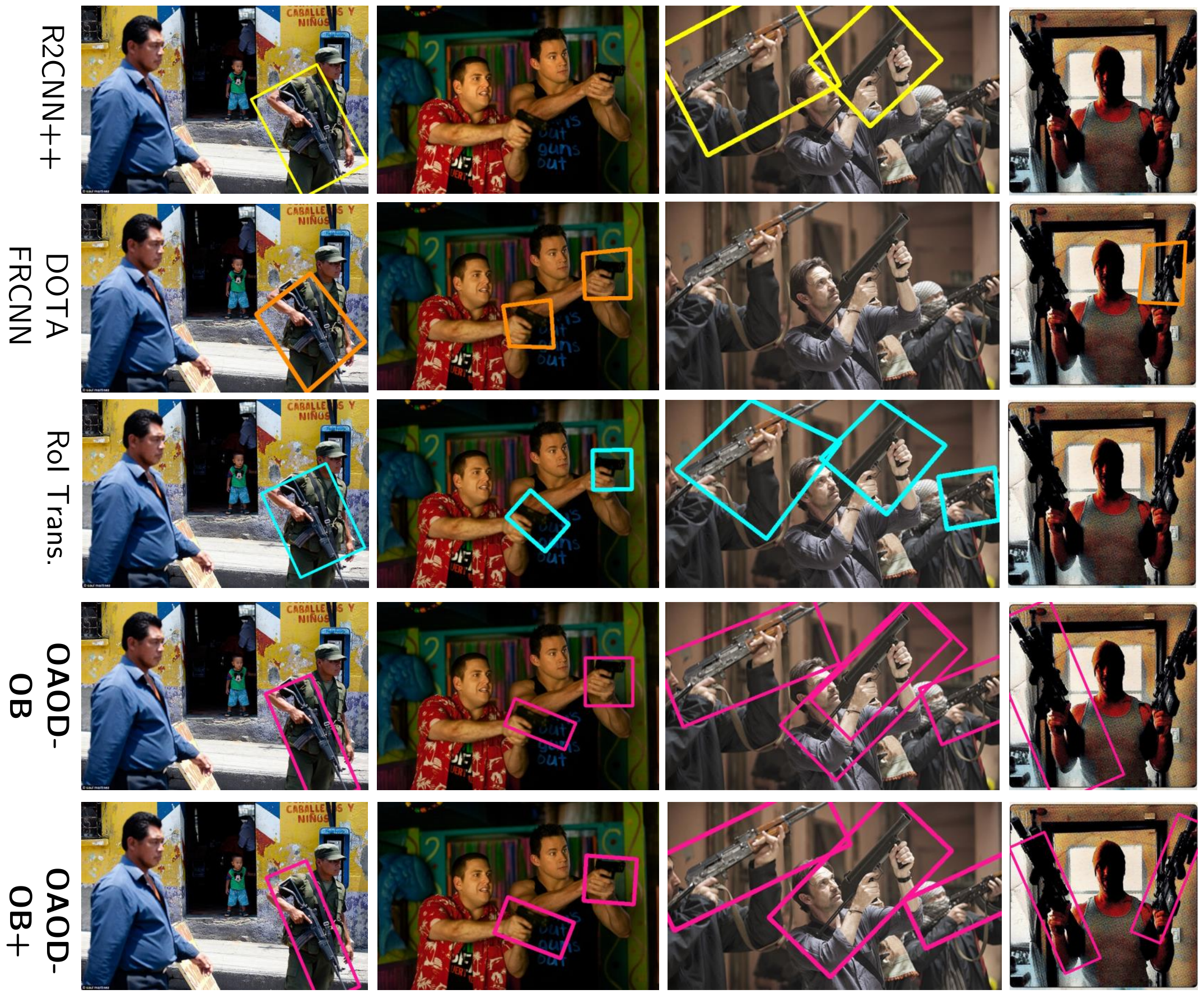}
        \captionsetup{justification=justified}
        \caption{\small{Qualitative comparison of proposed OAOD-OB \& OAOD-OB+ (\textcolor{magenta}{Magenta}) with existing oriented object detectors (R2CNN++: \textcolor{yellow}{Yellow}, DOTA-FRCNN: \textcolor{orange}{Orange} and RoI-Trans: \textcolor{cyan}{Cyan}). OAOD-OB+ avoids miss-detection, multiple detections and achieves more accurate localization. 
        }}
        \label{orientbbox}
    \end{center}
\end{figure*}

\subsubsection{Comparison with Oriented Object Detection Methods}
Most of these methods are trained using  OBB ground truth annotations, while such information is not available in the case of the ITUF dataset.  

In order to train the existing oriented bounding box detection methods (R2CNN++ \cite{ yang2018r2cnn}, DOTA FRCNN \cite{Xia_2018_CVPR}, RoI-Trans \cite{Ding_2019_CVPR}), we rotate the given AABB ground-truth with respective ground-truth orientation and create OBB.
OAOD overcomes the limitation of the unavailability of OBB ground-truth by leveraging the orientation information (stage-1), ORP with OARoI-Pooling and inverse transformation (stage-2).
Table. \ref{tab:oriented-comparison} shows the comparison of OAOD-OB \& OAOD-OB+ with the existing state-of-the-art oriented object detection algorithms. OAOD gives more stable results for different IoU values compared to other methods.
For a comprehensive evaluation, we have tested the proposed OAOD algorithm and the existing OBB detection methods with rotated OBB ($OBB_{rot}$) and annotated OBB ($OBB_{ann}$). The $OBB_{rot}$ are the rotated version of AABB and $OBB_{ann}$ are the manually tagged boxes (available for test set only).
More specifically, in the case of $OBB_{rot}$ and $OBB_{ann}$ at IoU=0.5, the proposed OAOD outperformed the existing state-of-the-art methods by a minimum of 5.8\% and 11.6\%  mAP, respectively. 

Fig \ref{orientbbox} shows the qualitative results of OAOD-OB \& OAOD-OB+ with the existing state-of-the-art oriented object detection methods. More Qualitative results of OAOD are provided in the supplementary material.

\begin{figure}[t]
    \begin{center}
    \captionsetup{justification=justified}
    \includegraphics[width=0.9\linewidth]{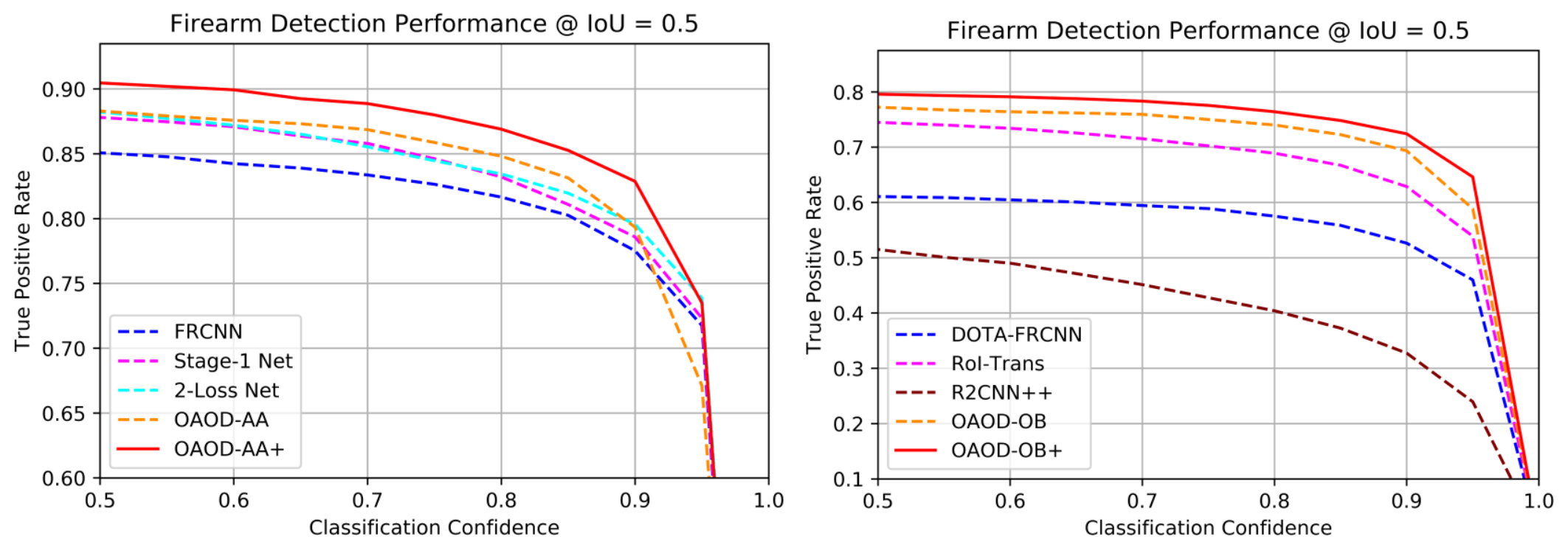}
    \caption{\small{OAOD's behavior across varying confidence levels. 
    \textbf{Left} figure shows AABB detection performance with ablation experiments. \textbf {Right} figure shows OBB detection performance using $OBB_{ann}$ over test set.}}
    \label{fig:OAODOBgraphs_r}
    \end{center}
\end{figure}

To compare the robustness and stability of the OAOD vs existing methods, we evaluate results at different confidence levels. As indicated in Fig. \ref{fig:OAODOBgraphs_r} (right), proposed OAOD-OB \& OAOD-OB+ algorithms remain stable and even at high confidence threshold, OAOD results less deteriorate  compared to the other methods \cite{Ding_2019_CVPR, Xia_2018_CVPR, yang2018r2cnn}. This could be attributed to the cascaded nature of OAOD and removal of the noisy background features by generating Oriented Proposals and performing Object Alighted RoI-Pooling.

\subsection{Ablation Study}
\label{sec:ablation}
Thorough ablation study is performed in order to evaluate different design choices including hyper-parameters  and OAOD precursor models (Fig. \ref{fig:OAODOBgraphs_r} (left)).

\noindent \textbf{Orientation Loss:} 
With other hyper-parameters in stage-1 loss (\ref{eq5}) fixed to 1  (following original FRCNN paper), we searched a wide range for the optimal value of $\beta$,
by training and validating on the ITUF validation-set chosen as 20\% of the training dataset.
$\beta=0.10$  has resulted in increased orientation accuracy as well as mean average precision (Table \ref{betaval}),  therefore $\beta=0.1$  is used for the rest of the experiments. 

\begin{table}[b]
\footnotesize
\centering
\renewcommand{\arraystretch}{1.2}
\tabcolsep=3.9pt\relax
\caption{\footnotesize{Orientation accuracy and mean average precision (stage 1) with varying values of  $\beta$ in \eqref{eq5} over the validation dataset. The \textcolor{red}{Red} represent high values and used in our experiments.}}
\begin{tabular}{cccccccc}
\hline
$\beta$ & 1   & 0.5 & 0.325 & 0.25  & 0.125          & 0.1   & 0.0625 \\ \hline
$\text{mAP}_{validation}$  & 51.5 & 62.9 & 66.6    & 72.5 & 71.9 &\textcolor{red}{74.8} & 72.5  \\ \hline
Accuracy & 84.4 & 83.5 & 84.3 & 84.2 & 83.9    & \textcolor{red}{84.7} & 82.9  \\ \hline

\end{tabular}

\label{betaval}
\end{table}

\noindent  \textcolor{black}{ \textbf{Orientation Regression Loss:} 
Similar to orientation loss in (\ref{eq5}), we search for the optimal value of of orientation regression loss scaling parameter $\eta$.
With other hyper-parameters in stage-1 loss (\ref{eq5}) fixed, $\alpha =1$, and $\gamma=1$ (following original values used in FRCNN \cite{ren2015faster}) and $\beta=0.1$, the optimal value of $\eta$ is searched by training on the training set (chosen as 80\% of the training dataset) and validating on the ITUF validation-set (chosen as 20\% of the training dataset).
$\eta=1.0~ \text{and}~ 0.5$  has shown comparative mAP (Table \ref{etaval}), however, $\eta=1.0$ have minimum absolute orientation error. Based on this observation, $\eta=1.0$ is used for the rest of the experiments.}

\begin{table}[t]
\footnotesize
\centering
\renewcommand{\arraystretch}{1.2}
\tabcolsep=3.9pt\relax
\caption{\footnotesize{Orientation absolute error and mean average precision (stage 1) with varying values of  $\eta$ in \eqref{eq5} over the validation dataset. The \textcolor{red}{Red} represent high mAP while the \textcolor{blue}{Blue} shows the values used in our experiments as absolute error is less in this case with comparable mAP.}}
\begin{tabular}{c|cccccc}
\hline
$\eta$     & 1.5   & 1.25 & 1.0 & 0.75  & 0.50          & 0.25 \\ \hline

$\text{mAP}_{validation}$  & 77.9 & 71.1  &\textcolor{blue}{78.2}    & 77.7  &\textcolor{red}{78.3}  & 77.6  \\ \hline
Absolute Error & 4.9 & 4.8 & \textcolor{blue}{4.4} & 4.8 & 4.6  & 4.9  \\ \hline
\end{tabular}

\label{etaval}
\end{table}


\noindent \textcolor{black}{\textbf{Orientation Classes Distribution:} 
To find the effective orientation classes $n_o$, We have validated orientation classes distribution, while having all the hyper-parameters ($\alpha=1,~ \beta=0.1,~ \text{and}~ \gamma=1$) in stage-1 fixed. The orientation classes distribution, respective mAP, orientation accuracies and orientation absolute errors (mean of the absolute differences between the predicted orientations and groud truth orientation) the  are shown in Table. \ref{theta_dist}. For $n_o = 4$, the orientation accuracies are higher but the respective mAP values are dropped significantly along with high orientation absolute error. This is due to the fact, that the model have to classify the orientation in less number of classes which is an easy task compared to more classes with the cost of decrease in mAP and increase in absolute orientation error. Similarly, for more orientation classes, e.g., $n_o = 12$, the orientation performance decreases due to very close mean angle values causing an increase in absolute orientation error, however a slight increase in mAP is reported. The similarly, we repeated the experiment for $n_o = 8$, resulting low absolute error with comparative mAP. 
Since orientation being the main component of the proposed OAOD algorithm, we choose $n_o = 8$ orientation classes with high mAP compared to $n_o = 4$ classes and with less orientation absolute error compared to $n_o = 12$ orientation classes as shown in blue color in Table \ref{theta_dist}. }

\begin{table}[t]
\footnotesize
\centering
\renewcommand{\arraystretch}{1.2}
\tabcolsep=3.9pt\relax
\caption{\footnotesize{Orientation accuracy, orientation absolute error and mean average precision (stage 1) with varying number of orientation classes in \eqref{eq5} over the validation set. The \textcolor{red}{Red} represent high mAP while the \textcolor{blue}{Blue} shows the values used in our experiments with smaller absolute error and comparable mAP.}}
\begin{tabular}{c|cccccc}
\hline
$\theta_n$   & 4 & 8  & 12 \\ \hline

$\text{mAP}_{validation}$ & 67.5   & 	\textcolor{blue}{74.8}   &  \textcolor{red}{76.6} \\ \hline
Accuracy & 91.1	  & 	\textcolor{blue}{84.7} & 	69.7 \\ \hline
Absolute Error & 10.6	 & 	\textcolor{blue}{3.8}	 & 	4.6 \\ \hline

\end{tabular}

\label{theta_dist}
\end{table}

\noindent \textbf{Stage-1 Net:} The stage-1 (\ref{fig:main-model}) of the proposed OAOD described in Sec. \ref{sec:method-phase-1} is also evaluated for axis-aligned firearm detection. It is noted that, incorporating the orientation information using multi-task learning has improved the performance over the baseline FRCNN by 3.1\% mAP (Table. \ref{tab:ioumap-ablation}). 

\begin{table}[t!]
\footnotesize
\centering
\renewcommand{\arraystretch}{1.3}
\tabcolsep=4.5pt\relax
\caption{\footnotesize{Comparison of OAOD with baseline (FRCNN), Stage-1 and 2-Loss Net for AABB task. Highest values shown in \textcolor{red}{Red}, $2^{nd}$ highest shown in  \textcolor{blue}{Blue}
}}
\begin{tabular}{c|ccccccc}
\hline
$IoU$                                                          & FRCNN   & Stage-1 Net & 2-Loss Net  & \begin{tabular}[c]{@{}c@{}}\textbf{OAOD-}\textbf{AA}\end{tabular} & \begin{tabular}[c]{@{}c@{}}\textbf{OAOD-}\textbf{AA+}\end{tabular} \\ \hline
0.4 & 88.9 & \textcolor{black}{88.9} & 88.8   & \text{\textcolor{blue}{89.2}} & \text{\textcolor{red}{89.9}}  \\ \hline
0.5  & 79.8 & 82.3 &   \textcolor{black}{82.9} &  \text{\textcolor{blue}{85.4}} & \text{\textcolor{red}{88.3}}  \\ \hline
0.6  & \textcolor{black}{68.1} & 66.1 &    65.9                  & \text{\textcolor{blue}{70.3}} & \text{\textcolor{red}{75.7}}  \\ \hline
\end{tabular}

\label{tab:ioumap-ablation}
\end{table}

\noindent \textbf{2-Loss Net:} In this experiment, the proposed cascaded model (Fig. \ref{fig:main-model}) is trained by using only AABB regression loss from stage-1 and  classification loss from stage-2 as $L_{2L} =  L^f_{2} + L^b_{1}$. The other losses are not used in this experiment.
The $RP_{1}$ are used for generating $ORP$ along with OARoI-Pooling, unlike the $RP_{2}$ used in OAOD. 
Compared to FRCNN  the 2-Loss Net improves the mAP by 3.8\%  (IoU=0.50) that shows the significance of our basic framework. 2-Loss net's mAP remains 6.1\% less than OAOD mAP that highlights the importance of the remaining losses.   

\begin{figure}[h!]
    \begin{center}
    \captionsetup{justification=justified}
    \includegraphics[width=0.6\linewidth]{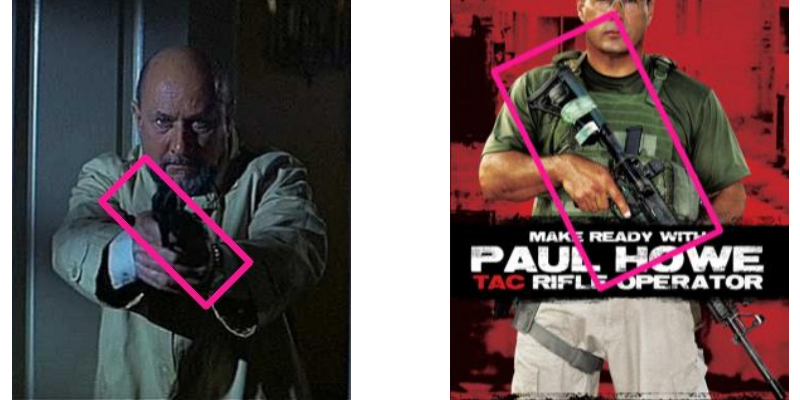}
    \caption{\small{Example OAOD failure instances: Left: extremely outward pointing object are hard to localize, Right: occlusion may cause localization errors. }}
    \vspace{-0.5cm}
    \label{fig:fail}
    \end{center}
\end{figure}

\noindent \textbf{Failure Cases:} Fig. \ref{fig:fail} shows two failure cases by proposed OAOD. The left image characterizes the case where the firearm is itself not detectable (due to viewing angle, pose, color, etc.). Only information is in the pose of the holder, we intend to explore connection between human pose and firearm localization in future work. 
In the right one, failure is only a partial one, as the main component has been localized with the correct orientation, while the barrel has been missed due to the occluded portion. One of the possible reasons could be the lack of such occluded objects in the training data.

\section{Conclusion}

Rising gun violence, and the use of firearms in both electronic media and social media, poses a challenge for the security, surveillance and multi-media content curation domains. However, there has been no concrete effort in the direction of the firearm detection problem. We counter it by, first, introducing a large challenging dataset of images containing firearms, named ITUF dataset which consists of $10973$ images, where all the firearm instances have been hand-annotated. Secondly, we propose a novel firearm detector using the oriented object detection technique for the firearm detection problem. Firearms, being thin (and many being elongated), and mostly held in oriented poses are perfect fit for the oriented object detection problem.

For this purpose, an Orientation Aware Object Detector (OAOD) architecture is designed that can detect tight oriented bounding boxes (OBB) while being trained in weakly supervised fashion using only axis-aligned bounding boxes (AABB) and orientation information. 
OAOD, is designed to be a multistage detector such that at the last stage features become independent of the object's orientation. 
Such a setup simplifies the task of classification and bounding box regression improving its accuracy.
To keep the number of anchor boxes small at RPN level, orientation is not associated with the region proposals.
Instead, an orientation prediction module is introduced, that predicts the orientation from every axis-aligned proposal classified as a firearm.
Predicted orientation is used in an oriented region proposal generation step that allows sampling of features around the region aligned with the orientation of the object inside the AABB predicted in the last stage.
We train OAOD, to detect OBB around the firearms, and classify them into two broad classes, guns and rifles. 
The experimental results (mAP: \textit{88.3} on $\text{AABB}$ \& mAP: \textit{77.5} for $\text{OBB}$) demonstrate the effectiveness and stability of the proposed method compared with the existing state-of-art methods.

\noindent \textbf{Acknowledgment:}

We greatly appreciate the assistance from Muhammad Faisal and Anza Shakeel in collecting and annotating the dataset, and Jason Chi for discussions and providing useful comments.







\bibliography{FirearmsDet_arxiv}




\end{document}